\documentclass[sigconf]{acmart}
\acmSubmissionID{5928}

\AtBeginDocument{%
  }

\usepackage{natbib}

\copyrightyear{2026}
\acmYear{2026}
\setcopyright{cc}
\setcctype{by}
\acmConference[IDC '26]{Proceedings of the 25th Interaction Design and Children Conference}{June 22--25, 2026}{Brighton, United Kingdom}
\acmBooktitle{Proceedings of the 25th Interaction Design and Children Conference (IDC '26), June 22--25, 2026, Brighton, United Kingdom}
\acmDOI{10.1145/3773077.3810976}
\acmISBN{979-8-4007-2283-7/2026/06}

\begin{document}

\title[Designing Robots to Support Parent-Child Connections]{Designing Robots to Support Parent-Child Connections: Opportunities Through Robot-Mediated Communication}

\author{Michael F. Xu}
\orcid{0009-0000-2632-2428}
\affiliation{%
  \institution{University of Wisconsin--Madison}
  \city{Madison}
  % \state{Wisconsin}
  \country{USA}
}
\email{michaelfxu@cs.wisc.edu}

% Tried to match what you had in the "Kept Alive, Bricked, Revived" paper
\author{Bengisu Cagiltay}
\authornote{Part of this work was conducted while the authors were affiliated with the University of Wisconsin--Madison.}
\orcid{0000-0001-7024-4957}
\affiliation{%
  \institution{Koç University}
  \city{Istanbul}
  \country{Türkiye}
}
\email{bcagiltay@ku.edu.tr}

\author{Yaxin Hu}
\orcid{0000-0003-4462-0140}
\affiliation{%
  \institution{University of Wisconsin--Madison}
  \city{Madison}
  % \state{Wisconsin}
  \country{USA}
}
\email{yaxin.hu@wisc.edu}

\author{Anjun Zhu}
\authornotemark[1]
\orcid{0009-0001-9583-8900}
\affiliation{%
  \institution{Simon Fraser University}
  \city{Burnaby}
  % \state{British Columbia}
  \country{Canada}
}
\email{aza99@sfu.ca}

\author{Bilge Mutlu}
\orcid{0000-0002-9456-1495}
\affiliation{%
  \institution{University of Wisconsin--Madison}
  \city{Madison}
  % \state{Wisconsin}
  \country{USA}
}
\email{bilge@cs.wisc.edu}

\begin{abstract}
The sense of \textit{family connectedness} may support positive outcomes including individual well-being, resilience, and healthy family functioning. However, as technologies advance, they often replace human-human interactions instead of nurturing them. In this work, we investigate how robot-facilitated communication tools might instead create new opportunities for family connection. We conducted two studies with families with children aged 5-12. We first explored the design space through in-home technology probe sessions with six families. These probes inspired us to explore two key interaction design dimensions: the \textit{robot's behavior strategy} (passive, reactive, proactive) and the \textit{mode of communication} (synchronous, asynchronous). We then conducted a laboratory study with 20 families to examine how the two dimensions shaped parent-child interaction and connection. Our findings characterize how parents and children appropriated robot-mediated exchanges, the tensions they experienced around initiative, timing, and privacy, and the opportunities they envisioned for supporting everyday connectedness.
\end{abstract}

\ccsdesc[500]{Human-centered computing~Interaction design process and methods}
\ccsdesc[300]{Computer systems organization~Robotics}
\ccsdesc[300]{Human-centered computing~Empirical studies in HCI}
\ccsdesc[100]{Human-centered computing~User studies}

\keywords{family connection; parent-child communication; social robot; technology probe; interaction design}
\begin{teaserfigure}
  \includegraphics[width=\textwidth]{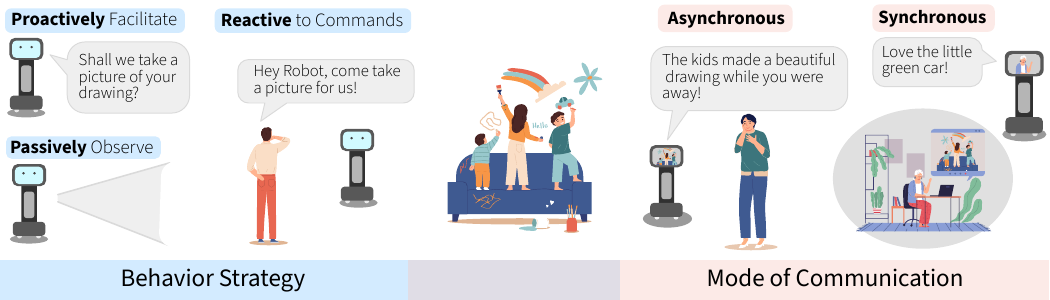}
  \caption{Motivated by the messy realities of everyday family life, we first explored the design space of robot-facilitated communication through iterative in-home technology probes. These exploratory sessions inspired us to further investigate two key dimensions: the \textit{robot's behavior strategy} (passive, reactive, proactive) and the \textit{mode of communication} (synchronous, asynchronous). We implemented a prototype highlighting these two dimensions and systematically evaluated it through a subsequent controlled user study. We derived design opportunities for robots to complement and strengthen family connection.}
  \Description[Robot facilitation strategies and communication modes]{Conceptual diagram showing two design dimensions for robot-mediated family communication. The left half presents behavior strategy, with three levels arranged from less to more active involvement: passively observe, proactively facilitate, and reactive to commands. Passively observe is illustrated by a robot watching from the side without speaking. Proactively facilitate is illustrated by a robot asking, ``Shall we take a picture of your drawing?'' Reactive to commands is illustrated by a person telling the robot, ``Hey Robot, come take a picture for us!'' The right half presents mode of communication, with two forms: asynchronous and synchronous. Asynchronous is illustrated by a parent speaking to the robot while the rest of the family is shown in another scene, with the robot relaying, ``The kids made a beautiful drawing while you were away!'' Synchronous is illustrated by a live interaction between a remote adult on a screen and the robot in the home, with the message, ``Love the little green car!'' A family scene in the center visually connects these dimensions, depicting a few family members engaged in a shared activity on a couch.}
  \label{fig:teaser}
\end{teaserfigure}

\maketitle

\section{Introduction}
% % broader social connection / loneliness --> family
% Social connectedness is increasingly recognized as a public health concern. In recent years, reports have documented rising levels of loneliness and social isolation, with nearly one in three adults in the United States reporting feeling lonely at least once a week~\cite{apa24poll, apa25poll}. In 2023, the U.S. Surgeon General Vivek Murthy has characterized this as an ``epidemic of loneliness and isolation,'' noting its harmful effects on health and well-being comparable to other major public health risks~\cite{office2023our}. Among the many contexts in which people seek connection, \textit{families} are often the first and most enduring. In his own recount about escaping the pain of loneliness, Vivek described how daily calls with his parents and sister were instrumental~\cite{nytLoneliness}, underscoring the importance of strong family connectedness to individuals, families, and society alike.

% strong connections within a family is important / desirable
\textit{Family connectedness} is often described as a family's sense of psychological closeness~\cite{resnick1993impact, crespo2010relationships}, typically characterized by open communication, quality time shared together, and mutual support during challenging times. Connections among family members play a critical role in shaping individual well-being and the overall functioning of family systems; connected families foster positive relationships that promote adolescent mental health and adjustment~\cite{woodman2018young}, demonstrate greater resilience in the face of hardship~\cite{black2008conceptual}, and experience healthier outcomes overall~\cite{parvizy2009qualitative}. Beyond these benefits, families also exhibit a strong and natural desire to stay connected~\cite{neustaedter2006interpersonal}.

Digital technologies have played a mixed role in fostering family connectedness. On one hand, phones, tablets, and video calls provide more ways for remote family members to connect. More experimental devices have also sought to reduce the effort required to maintain such connections, such as wearable cameras~\cite{hodges2006sensecam}, always-on recording systems~\cite{heshmat2017autobiographical}, family media spaces \cite{judge2010family}, and even explorations into multi-sensory design spaces~\cite{shakeri2023sensing}. However, as technology becomes increasingly pervasive in our lives, it is unclear whether they are in fact fostering, or replacing, human-human connections. As \citet{10.5555/1972496} described in her book \textit{Alone Together}, ``\textit{we expect more from technology and less from each other.}''

Robots, while sharing some of the caveats of other technologies, introduce new design possibilities. Instead of replacing communication among family members, recent works have demonstrated the potential of robots facilitating and enriching such communication, for co-located family members such as parent and child~\cite{ho2023designing, ho2024s, chen2025social}, remote family members~\cite{seo2024feel, moyle2020telepresence}, and inter-generational communications~\cite{mo2017enhance,joshi2019robots}, highlighting the potential for robots to support everyday family connection in ways that complement, rather than replace, existing human-human interactions, \textit{if designed properly}.

Building on this promise, we explore the design space of robot-mediated parent-child communication, focusing on when and how such systems may support connection-making and what trade-offs they introduce in family life (Figure~\ref{fig:teaser}). Here, \textit{communication} is characterized by the transmission or expression of ideas, emotions, and feelings, as well as the exchange of information~\cite{mw:communication, runcan2012role}. Given the messy, complex realities of everyday family life, and the lack of prior work outlining this design space, we began with an iterative design process through exploratory in-home technology probe sessions to understand how families might envision a robot participating in everyday activities to facilitate connection-making. These sessions revealed socio-technical insights into how a robot might facilitate communication between parents and children during key connective moments, as well as two key interaction design dimensions: behavior strategy (passive, reactive, proactive) and mode of communication (synchronous, asynchronous).

% This technology probe inspired the central idea of a robot facilitating connection and communication between parents and children during key moments within the home as well as two key design dimensions: the robot's behavior strategy (passive, reactive, proactive) and the mode of connection it facilitates (synchronous, asynchronous). 

Building on these insights, we adopted a mixed-design approach and conducted a follow-up in-lab user study with 20 additional families. Through structured interactions with both a mobile telepresence robot and a tablet-based communication platform, this study enabled us to systematically examine how interaction paradigms and the two design dimensions identified in the prior study shaped family-robot interactions and intra-family communication. In particular, the in-lab setting allowed us to identify distinct advantages, limitations, and use scenarios for synchronous and asynchronous communication modes, as well as to probe families' expectations and concerns around the three robot behavior strategies and across different age groups of children.

% Building on these insights, we adopted a mixed-design approach and conducted a follow-up in-lab user study with 20 additional families to systematically examine how the interaction paradigm and the two design dimensions that emerged from the previous study shaped family-robot interactions and intra-family communication. Families generally recognized the potential of a mobile social robot to facilitate connection-making and expressed appreciation for both asynchronous and synchronous modes of interaction, noting that each offered distinct advantages and was useful in different scenarios and for different age groups. While participants voiced reservations about proactive robot behaviors, they also acknowledged that such behaviors could be valuable in specific contexts.

% Taken together, the two studies extend our understanding of how robot strategies and interactions may shape family connectedness. We report on the design and execution of both studies, synthesize insights from observations and interviews, and derive design implications for future robots intended to support connection-making within families.

In sum, our work makes the following contributions:
\begin{enumerate}
    \item \textit{Design Space:} We explored and examined the design space for robots that support family communication and connectedness;
    \item \textit{Artifact:} We iteratively designed a prototype mobile robot system capable of facilitating various modes of parent-child interactions;
    \item \textit{Design Recommendations:} We analyzed families' feedback and distilled implications to guide the development of future in-home robots intended to support family connections.
\end{enumerate}

\section{Background}
The benefits of strong connections among family members are well established, ranging from improved well-being to greater resilience and healthier family functioning~\cite{woodman2018young, black2008conceptual, parvizy2009qualitative}. What remains less clear is under what conditions, and with what trade-offs, technologies can meaningfully support connection in everyday life. This question is especially important in families, where communication unfolds amid varying routines, availability, and privacy expectations.

\subsection{Technology-Facilitated Connection}
Families have long turned to technologies to maintain a sense of connection. Communication tools such as video calls, messaging platforms, and shared family apps provide straightforward ways to stay in touch. \citet{judge2010family} explored the use of a ``Family Window,'' a media space featuring videos of each other between two households, and found that increased awareness of each other's availability created more opportunities to share everyday experiences, which in turn strengthened feelings of connectedness, corroborating earlier work on the importance of \textit{interpersonal awareness}~\cite{neustaedter2006interpersonal}. More recently, researchers have explored the potential of rich media such as videos and photos to support collective memory amongst family members and beyond. For example, \citet{heshmat2017autobiographical} conducted an autobiographical design study with an always-on video-taping system, and \citet{uriu2009caraclock} prototyped a photo-viewing device, both highlighting how people valued technologies that supported shared memories and experiences.

Prior work has also investigated the role of technology focusing on connections between parents and their children. In an investigation of digital media use and its impact on parent-child connection, \citet{padilla2012getting} found that watching TV and movies or playing video games \textit{together} were associated with higher levels of family connection, while higher levels of engagement on social networking sites were associated with lower levels of perceived connection. As an example of a tangible interface, \citet{golsteijn2013facilitating} designed and implemented an interactive digital photo ``album,'' the \textit{Cueb}, through which parents and their teenage children are prompted to exchange stories and explore experiences. Comparing a baseline version and a ``smarter'' iteration of the Cueb, the authors reported that parents and children were more willing to communicate their experiences when the presented photos were related or covered shared experiences between the parent and child, highlighting the value of intentionally curated memory cues.

Although prior work has demonstrated the value that technologies can bring in facilitating intra-family connection, important challenges remain. In a systematic review of technologies supporting parent-child relationships, \citet{shin2021designing} identified two recurring challenges: differences in parents' and children's expectations around communication, and the busy schedules of parents, which often give rise to complex emotional experiences surrounding parenting. At the same time, the technologies explored to date have been predominantly disembodied and positioned as neutral communication tools or media. Building on this prior work, we explore how integrating a social robot --- potentially through its embodiment, social presence, and capacity to respond to family contexts --- can reshape \textit{when}, \textit{how}, and \textit{for whom} technology-mediated communication is experienced as meaningful within family life.

% % % https://dl.acm.org/doi/10.1145/3520495.3520515
% % Using mobile devices could potentially lead to conflict In Parents' relationships

% % https://www.sciencedirect.com/science/article/pii/S1876201821000514
% Offline family communication bonding events are negatively affected by technology use

% % https://onlinelibrary.wiley.com/doi/full/10.1111/j.1741-3729.2012.00710.x
% Family media activities such as coviewing videos can strengthen family connection, whereas social media networking is linked to lower connection.

% % https://dl.acm.org/doi/10.1145/2858036.2858157
% Properly using smartphones can help build children's independence and and reduce parents' frustration levels.

% % https://onlinelibrary.wiley.com/doi/epdf/10.1111/fare.12960
% Social media as a tool can facilitate the connection between children and deployed parents

\subsection{Robots in Families}
% Robots have been increasingly introduced and deployed in family settings. We first take a high-level view of the context they are embedded in and the roles they take on, and then review how they have been designed and utilized to facilitate interaction and communications within families.

Social robots have been envisioned to play a variety of roles in family setting. For example, prior work has introduced robots as companions and play partners for children~\cite{abe2018chicaro,mchugh2021unusual,levinson2022living}, as storytelling or learning companions~\cite{chen2022designing, michaelis2018reading}, as assistants for childcare~\cite{lee2022can}. These roles highlight families' interest in engaging with robots in varied ways, and show how different family members may hold different expectations for what a robot can or should do~\cite{cagiltay2020investigating}.

More importantly, the wide range of roles and applications of these robots also underscore the opportunities and potentials for robots to facilitate connection-making, as they become more and more embedded in all the various aspects of family lives. Recent works have explored the utility of robots in enriching family interactions and communications. For example, \citet{chen2025social} studied long-term dyadic interactions of more than 70 parent-child pairs, and showed enhanced quality of parent-child conversations due to the robot's active participation. \citet{ho2023designing} examined how a social robot might support parent-child conversations around math, finding that all nine participating parents felt the robot offered them new ideas for discussion and ways to integrate those conversations into everyday interactions with their children. Similarly, \citet{kim2022can} compared family-robot interactions to child-robot dyads, showing that when families played together with a robot, both verbal and physical engagement increased. In addition to supporting co-located family members, research has shown that robots can also facilitate connections with remote relatives and across generations. For example, \citet{seo2024feel} showed that robots can support remote family interactions by leveraging its mobility to engage in multi-party family communication. In nursing home settings, \citet{moyle2020telepresence} found that family carers experienced a stronger sense of connectedness when communicating through telepresence robots, noting that the robots supported longer conversations and enhanced social connection with their relatives.

In short, work in human-computer interaction has shown that specific designs can help families stay connected, but robots offer broader capabilities: mobility, interactivity, and an ability to embed themselves in daily routines, as demonstrated by the array of roles robots already play in the home. These underscore robots' potential to go further by facilitating connection across diverse family contexts, dynamics, and structures. While prior studies demonstrate that robots have the potential to mediate communication, they typically focus on specific use cases. In this work, we take a broader view, first conducting exploratory in-home design sessions to investigate the design space, and then through more controlled laboratory studies, systematically analyze the opportunities and tensions in robot-facilitated family connectedness.

% To comprehensively explore the problem and solution space, we began with an exploratory in-home technology probe study employing a Wizard-of-Oz (WoZ) approach. In the Formative Study, six families engaged in a regular household activity while the robot roamed around to capture moments of interest. Through this exploratory study, we identified key dimensions of the design space: the robot behavioral strategy in terms of proactive and reactive interactions, and the mode of communication in terms of synchronous and asynchronous connections. Based on these insights, we designed a prototype that supported these interactions (the artifact),  systematically evaluated them in a user study with an additional 20 families, and derived design recommendations for designing robots to facilitate family connections. This research was approved by the authors' Institutional Review Board. 

\subsection{Research through Design}
The initial exploratory phase of this study is grounded in a \textit{Research through Design} (RtD) approach, which treats design practice as a way of generating knowledge through making and reflecting on artifacts, rather than producing finalized solutions~\cite{zimmerman2007research, gaver2012should}. Within this broader RtD framing, we use \textit{Technology Probes} as a concrete method to gather situated feedback and guide iterative changes to our designs. Technology probes are intentionally simple and open-ended systems deployed \textit{in situ} to reveal how people use, adapt, and sometimes struggle with technology in everyday settings, particularly in complex domestic contexts~\cite{hutchinson2003technology}. By embedding probe-like artifacts within an RtD process, our study explores the design space in real-world contexts, not as a final evaluation, but as a way to learn across iterations and to surface key design dimensions and tensions for embodied technologies that support family connection.

%BENGISU'S RECOMMENDED RE-TITLING
\section{Technology Probe Design}
\label{sec:design}

% \subsection{Robot-Facilitated Communication Design}
% describe/move the "design concept" related content here.
% \label{sec:design_concept}
Given the potential benefits of improved communication and connection within families, and the various drawbacks current solutions experience, our design concept envisions a social robot that serves as a facilitator of family communication, helping parents and children share moments and experiences, stay engaged with and updated about each other, and most importantly, maintain a sense of closeness. Rather than replacing existing practices, the robot's interaction is designed with the intention to weave into the everyday life of families, capturing and sharing meaningful snippets, prompting interactions at opportune times, and offering subtle reminders that encourage communication.

Towards that end, we adopted a Research through Design approach~\cite{zimmerman2007research}. We initially envisioned futures shaped by novel robot designs and iteratively explored the design space (Figure~\ref{fig:method}). In this small-scale, exploratory study phase, we emphasize the \textit{inspirations} drawn from this design process, how literature from the interaction design and communication fields guides our interpretation of these empirical inspirations, and how together they inform the design space of robot-facilitated communication. Through this process, we developed a final artifact that we systematically evaluate in follow-up structured laboratory sessions (Section~\ref{sec:userstudy}).

\begin{table*}[t]
\centering
\caption{Demographic Information of Families who Participated in the Technology Probes}
\label{tab:study1-families}

\begin{tabular*}{\textwidth}{@{\extracolsep{\fill}} c c l l l l l @{}}
\toprule
\textbf{Family Code} & \textbf{Artifact} & \textbf{Parents} & \textbf{Children} & \textbf{Ethnicity} & \textbf{Races} & \textbf{Income} \\
\midrule
F1 & \#1 & 39F, 42M & 6M, 4M, 3F & PNR & PNR & \$150k--\$200k \\
F2 & \#2 & 88M, 37F & 9F, 7M & Non-Hispanic/Latino & Asian, White & \$50k--\$75k \\
F3 & \#2 & 39F, 39M & 10M, 13M & Non-Hispanic/Latino & White & \$100k--\$150k \\
F4 & \#2 & 53M, 48F & 12M, 9F & Non-Hispanic/Latino & Asian, White & \$100k--\$150k \\
F5 & \#2 & 43F, 43M & 11F, 7F & Mixed Hispanic/Latino Status & White & \$150k--\$200k \\
F6 & \#2 & 42F, 43M & 13F, 10F & Non-Hispanic/Latino & Asian & \$100k--\$150k \\
\bottomrule
\end{tabular*}

\vspace{2pt}
\begin{flushleft}
\footnotesize \textit{Note.} PNR = Prefer not to respond; \$100k = \$100{,}000.
\end{flushleft}
\end{table*}

\subsection{Probe Procedure}
% describe the study design, participant recruitment, ethics etc.
Given the vast and open-ended nature of our design concept, we began the investigation with a series of exploratory, in-home technology probes, employing a Wizard-of-Oz (WoZ) approach. We recruited families with at least one child between the ages of 5 and 12, through the university's research recruitment mailing list. In total, six families and 13 children participated in the probes. Participant demographics are provided in Table~\ref{tab:study1-families}. Prior to participating in any research activities, all adult participants completed an informed consent process, and all minors completed an oral assent process. The study lasted around two hours, and participating families received $\$50$ USD for completing the study. This research was approved by the authors' Institutional Review Board. 

\subsubsection{Home Visits}
The home visits consisted of three main phases.

\textit{Introduction and Setup:} Two experimenters visited the family's home together. Once the consent and assent processes were complete, one experimenter gave the family a brief introduction of the study procedure, while the other set up the system and equipment. This process took around 20 minutes.

\textit{Main Activity:} The families were prompted to select an activity that they would do for the next hour, and together with the experimenter, agree on an area within which the robot would navigate around. The experimenter provided some activity options, such as a kit for building a marble run, but families were also welcome to carry out an activity of their own. Once the family was ready, the experimenters stepped out of the home, and proceeded with the WoZ operations from a vehicle outside, relying solely on the robot's sensors with the exception of video streams from two mounted cameras. Families were not aware of the WoZ operation, but were aware of being recorded and observed. Families continued with their choice of activity for around 60 minutes, interacting with the robot as they desire. At the same time, a 360° camera\footnote{\url{https://www.insta360.com/}} was attached on top of the robot to capture the on-going activities. In later iterations of the research (detailed further in Section\ref{sec:design_techprobe}), the experimenters initiated interactions on-the-fly through the robot as opportune moments arise. For example, if the family were building something, the robot might ask if they want to take a picture with the creation at selective milestones.

\textit{Semi-Structured Interview:} After an hour of family-robot activities, the experimenters returned to the home, and conducted a semi-structured interview with the family. The interviews largely covered three components: (1) user experiences of the robot's presence and interactions, (2) a retrospective review of selected video clips captured by the robot, and (3) perception and reflections about the robot, how it could facilitate activities and support connections within families. %For example, we probed family preferences regarding the robot's behavior and interactivity, social roles, as well as their thoughts and concerns for the various existing or potential features the robot have or could have. 
These interviews lasted between 30 and 60 minutes.

\subsection{Iterative Interaction Design Artifacts}
% Describe the robot and resources you used as artifacts. Describe how they iteratively evolved.

% \todo{Discuss how we want to restructure this. The purpose? (1) How we explored the design space and arrived at the two dimensions; (2) an intro to the prototype used for the main study. (2) can be made much shorter if we do it separately. (1) is something I'm struggling with a bit (i.e. how much value does the exploration process add to the paper, and how much justification is needed for our choice of the 2 dimensions explored in the main study)}

%Bengisu's suggestions: perhaps move some parts from 3.2 to this section. E.g., give an overview of the RtD and Figure~\ref{fig:method} here, since that's the more holistic method that captures all stages (in-home iterative design and tech probe + Artifact + in-lab evaluation)
%then in 3.2 just talk about the in-home iterative design sessions, and how the tech probe study was conducted. 
% On another thought, Maybe I didn't quite get the separation of the Design Concept and Iterative Design Process sections. Perhaps that signals that they can be merged, and instead of subsections, you could use italics and bold do differentiate between concepts. 
%SUMMARY: This section needs some structural reformatting for flow and clarity of the argument.

% \subsection{Iterative Design Process through Technology Probes}
\label{sec:design_techprobe}

Following a Research through Design process, we iteratively synthesized families' feedback from our home visits and incrementally built the system to be  used in the laboratory study (Section \ref{sec:userstudy}). More broadly, this artifact development process aligns with prior family-robot design work that has used participatory design, lived technology experiences, and in-home deployments to iteratively shape domestic robot platforms for family life~\cite{chen2022designing, he2025developing,cagiltay2020investigating,michaelis2023off}. Below we describe the iterative evolution of three robot systems and insights from our RtD process.%evolution between three versions of the robot system created as a result of this RtD process, and the insights we've gained along the way.

% \paragraph{Base System} In addition to a 360° camera we attached on top of the robot, we also setup multiple recording devices in the activity area, both for data collection and to facilitate the WoZ operations.  The experimenters, who were also the wizards, had access to the video and audio feed. For our pilot session with F1, we used the Temi v1 robot~\cite{TemiURL}, while for F2-F6, we used the Double 3 robot~\cite{DoubleURL}. The experimenters were able to control the robot movements, and for the Double 3 robot, the experimenters were also able to control the robot's speech, along with the text and/or an emoji to be displayed on the screen. Moreover, with the Double 3 robot, users were presented with two buttons that they could interact with, one for recording a video clip, and the other for taking a picture. Example example snapshots of the robot screen, along with some further details about the study, are provided in Supplementary Materials\footnote{Supplementary Materials are accessible here: \url{https://osf.io/re3uh/?view_only=bc11cd33fd19480c8891a9c745cdf026}.}.

\begin{figure*}[tb]
  \centering
  \includegraphics[width=\linewidth]{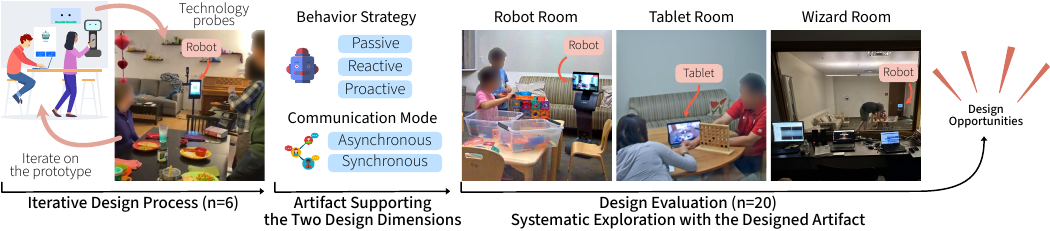}
  \caption{\textbf{Overview of our Research through Design process}. Our iterative design process started with in-home technology probe sessions where we incrementally enriched the prototype across sessions. Inspired by the technology probes and guided by prior literature, we identify key dimensions of the design space: the \textit{robot's behavioral strategy} (proactive, reactive, passive), and the \textit{mode of communication} (synchronous, asynchronous). We designed and implemented a system that supported these interactions (\textit{i.e.}, the artifact). We then conducted a user study to systematically examine these two factors, and derived design opportunities. The robot shown in the technology probe corresponds to \textit{Artifact \#2}, and the robot shown in the evaluation study corresponds to \textit{Artifact \#3}.}
  \Description[Study overview from probe design to evaluation]{Overview diagram of the study process. The figure is organized from left to right as a pipeline with three main stages. The first stage, iterative design process with six families, shows an early technology probe session and an in-home robot deployment connected by curved arrows to indicate iteration on the prototype. The second stage presents the designed artifact and highlights the two design dimensions explored in the study: behavior strategy, with passive, reactive, and proactive options, and communication mode, with asynchronous and synchronous options. The third stage, design evaluation with twenty families, shows the experimental setup across three coordinated spaces: the robot room, where children interact with the robot in the home; the tablet room, where another family member participates through a tablet; and the wizard room, where a researcher operates the system behind the scenes. A final arrow points to design opportunities, indicating that insights from the evaluation informed broader implications for design.}
  \label{fig:method}
  \vspace{-12pt}
\end{figure*}

\subsubsection{Artifact \#1: Passive Robot}
Prior research highlights the importance of shared \textit{moments} and everyday \textit{experiences} in sustaining family connectedness~\cite{dalsgaard2007ekiss, brush2008sparcs, romero2007connecting}. Guided by this literature, and consistent with our design vision, 
% vision for a robot that would help shape and nurture a sense of connectedness within families, 
the initial version of our artifact centered on how a robot might capture and surface such experiences within family life. Specifically, in this iteration, the robot adopted a fully passive role, navigating around the home and passively recording videos through the attached 360° camera. The base robot platform we used for \textit{Artifact \#1} was the Temi V1 robot\footnote{\url{https://www.robotemi.com/}}, which features built-in telepresence-style console for ease of control. 
% A 360° camera was attached on top of the robot to capture the on-going activities, and the footage was later reviewed with families to facilitate the semi-structured interviews.

\paragraph{Preliminary Insights}
We conducted the technology probe with F1. While the first family confirmed their preference for the robot as a passive observer for most of the time, they also recognized the potential value of both robot-initiated and family-initiated interactions. In particular, the father highlighted the usefulness of the robot capturing predictable events, suggesting that a robot able to respond to direct commands or proactively recognize significant upcoming moments could be beneficial: ``\textit{imagine we're about to blow out the candles, [...] and [we say] `film this!' }'

From a theoretical perspective, the desire for system intelligence integrated with user control is consistent with prior literature in interaction design. While passive capture supports the documentation of memory and special moments without intervention, the request for direct commands aligns with theories of agency and autonomy~\cite{deci2008self,beer2014toward}, indicating how the various levels of interactivity and autonomy may impact user's experience, perception and acceptance, among other factors. Conversely, the desire for the robot to ``recognize'' moments resonates with frameworks of implicit interactions~\cite{ju2008design}, which posit that intelligent systems should be capable of shifting initiative to support users when their attention is occupied elsewhere, moving between the background and the foreground as needed. Moreover, this integration of human control over automated behaviors is also consistent with guidelines from the broader Human-Centered AI framework~\cite{shneiderman2020human}. Motivated by these theoretical perspectives and our pilot findings, we expanded the design of our artifact to explore this spectrum of behavior strategies: passive, reactive, and proactive.

\subsubsection{Artifact \#2: Robot with Behavior Strategies}
Specifically, we designed and implemented Artifact \#2 that supported three interaction paradigms. With \textit{passive} behavior, the robot roams around and makes video recordings (as in the pilot session). With the \textit{reactive} behavior, the users are able to interact with the robot via either on-screen buttons (\textit{e.g.}, Start Recording, and Take Picture), or via voice commands (facilitated by Wizard-of-Oz (WoZ) operations). With \textit{proactive} behaviors, the robot initiates interactions, such as prompting the family to take pictures. \textit{Artifact \#2} was implemented on a Double 3 robot\footnote{\url{https://www.doublerobotics.com/double3.html}}, for its higher camera resolution, and more flexible capturing angles (\textit{e.g.}, telescoping poles and adjustable cameras). Example snapshots of the robot screen, along with further details about system setup, are provided in Supplementary Materials\footnote{Supplementary Materials are accessible here: \url{https://osf.io/re3uh/}.}.

\paragraph{Preliminary Insights}
Using variations of Artifact \#2, we conducted the technology probe sessions with five additional families. We confirmed the previous insight about \textit{appropriate robot behavior strategies} and identified a new dimension: \textit{the potential of supporting synchronous communication among family members}. During the interviews, families either spontaneously proposed telepresence-style communication as a potential use case for connection-making, or responded positively when asked by the interviewer. The mother in F6 reflected on the feature, ``\textit{I like that idea. That's great. [...] for working parents, [...] [if] I want to talk to her, like cheer her [up] [...] Good function to have!}'' The positives are not without caveats, though. For example, the father of F3 described how he envisioned using it: ``\textit{There are specific people that we give that ability to, and we are the ones that initiate: alright we're going to turn on the `grandpa robot', and we'll tell grandpa, `hey you can see our Christmas.'} '' These considerations also reflect existing work in the computer-mediated communication literature, where the synchronicity of the communication has been studied in various contexts, and shown to be an important factor of consideration both in terms of the application and use cases~\cite{chan2011shyness,park2015can,watts2016synchronous,jourdan2006perceived}.

In terms of robot behavior strategy, families expressed a general preference for a passive robot that remained in the background most of the time. At the same time, they discussed considerations for both reactive behavior (\textit{i.e.}, responding to user commands) and proactive behavior (\textit{i.e.}, autonomously initiating actions), though there was less agreement about which strategy was most suitable in which scenarios, or how such strategies should be implemented. These nuanced feedback motivated the design and implementation of the final artifact, and the user study that ensued.

\begin{figure*}[tb]
  \centering
  \includegraphics[width=.9\linewidth]{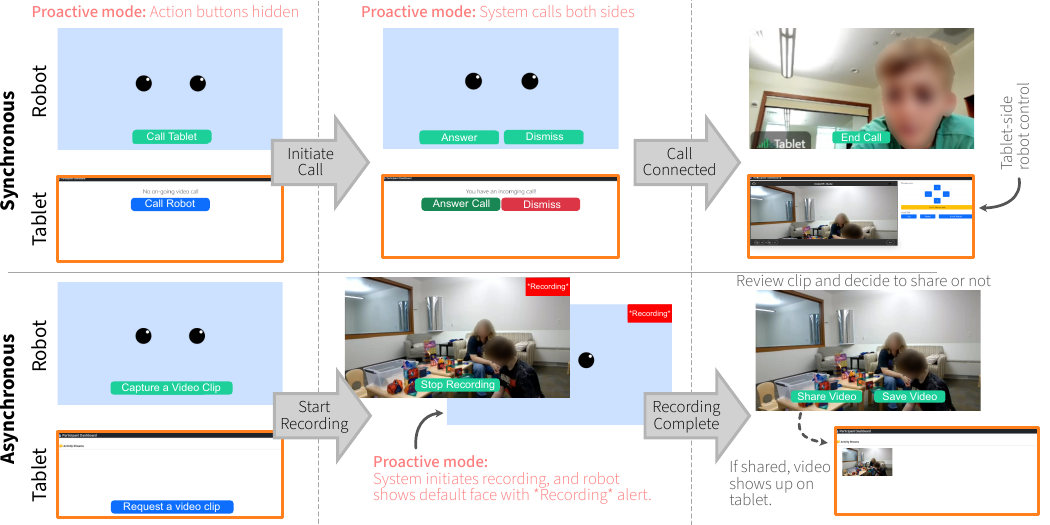}
  \caption{Illustration of the screens seen by the participants in the various scenarios in both conditions of the user study. Main UI differences between proactive and react mode are highlighted. }
  \Description[Interface flow for synchronous and asynchronous communication]{Interface walkthrough showing how the system supports two communication modes, synchronous and asynchronous, across robot and tablet displays. The figure is divided horizontally into a top synchronous section and a bottom asynchronous section, with each section showing a sequence from left to right. In the synchronous flow, the robot screen and tablet screen first show buttons to start a live call. An arrow labeled initiate call leads to a second state where both sides receive an incoming call interface with answer and dismiss options. A second arrow labeled call connected leads to the active video call, where the tablet-side user appears on the robot display and the robot-side camera feed appears on the tablet. A small control panel on the tablet allows tablet-side robot control. Notes indicate that in proactive mode the system can hide action buttons initially and automatically call both sides. In the asynchronous flow, the robot and tablet first show options to capture or request a video clip. An arrow labeled start recording leads to a recording state in which the robot-side scene is captured and a recording alert is shown. A second arrow labeled recording complete leads to a review state where the recorded clip can be watched and either shared or saved. A note indicates that if the clip is shared, a thumbnail appears on the tablet for later viewing.}
  \label{fig:screens}
\end{figure*}

\subsubsection{Artifact \#3: Final Prototype with Modes of Communication}
% The results from the technology probe sessions with Artifact \#2 identified two key design dimensions for a robot intended for facilitating family connectedness:  the robot's \textit{behavior strategy} (passive, reactive, proactive) and the \textit{mode of connection} it facilitates (synchronous, asynchronous).
Based on these insights, we designed and implemented the final prototype, a system that included a tablet and a mobile robot, supporting both synchronous and asynchronous conditions, and both proactive and reactive strategies.

The synchronous and asynchronous conditions were defined by robot-facilitated video calls, and the sharing of videos captured by the robot, respectively. In the \textbf{synchronous} condition, the family members connected with each other through a video call. Users have the option to accept or decline the call, and if accepted, the tablet's UI can be used to control and drive the robot. For the \textbf{asynchronous} condition, the robot took 30-second video clips, and shared it with the tablet-side user. The users have the option to decline sharing of the clip. 
%Users have the option to decline the call (synchronous) or  not share the recorded video clip (asynchronous). 

Illustrations of both the robot's and tablet's screens are shown in Figure~\ref{fig:screens}. Proactive and reactive behaviors were defined by the inclusion or exclusion of certain action buttons. For example, in proactive mode, the system (\textit{i.e.}, the wizard) initiates the interactions at predetermined times. For the synchronous condition, it simulates a call to both the tablets and the robot, which would connect if both sides answer. In the asynchronous condition, the robot moves towards the activity area and autonomously takes a video clip. Due to hardware issues with the Double 3 robot (\textit{Artifact \#2}), we've implemented our interaction design for \textit{Artifact \#3} on a Temi V3 robot\footnote{\url{https://www.robotemi.com/product/temi/}} -- an upgraded hardware compared to \textit{Artifact \#1}.

\section{User Study}
\label{sec:userstudy}

% Findings from the Formative Study offered a preliminary view of key design considerations, which we systematically examine in the User Study. 
% With respect to the robot's behavioral strategy, both proactive and reactive approaches may have distinct advantages and drawbacks. 
% Although the Formative Study artifact primarily supported asynchronous intra-household communication through the capturing and sharing of experiences and moments, families reflected on the potential importance of synchronous communication in the interviews.
Given the complexity of the design space described in Section \ref{sec:design}, we consolidated the design insights and conducted a user study using the robot prototype from Artifact \#3. This study followed a structured approach to evaluate families' use patterns and took place in a controlled laboratory environment, staged as a living room and play-room.

%we recognized the need for a more structured and systematic approach to evaluating families' use patterns and preferences.
%better understand the relevant factors and translate them into design guidelines. 
% To this end, we designed and conducted a user study in a controlled laboratory environment, staged as a living room and play-room, following a structured study protocol.

\subsection{Participants}
We recruited families with at least one child between the ages of five and twelve. Other children in the household were also welcome to participate, even if they fell outside this age range. Specifically, the eligibility inclusion criteria for the parents were: Fluent in English, 18 years or older, has at least one child aged 5-12 willing to participate.
% In the Formative Study, we also screened the participants based on whether they lived within <a specific geographic area, removed for blind review>.
In total, 20 families (27 children) participated in the user study, with demographics information provided in Table~\ref{tab:study2-families}. Prior to participating in any research activities, all adult participants completed an informed consent process, and all minors completed an oral assent process.

\subsection{Study Design}
% Given our goal to systematically explore two factors --- \textit{proactive and reactive strategies}, and \textit{synchronous and asynchronous communications} --- 
We adopted a mixed design approach where 10 families experienced the \textit{synchronous} condition, while the other 10 experienced the \textit{asynchronous} condition. Each family experienced one communication mode (\textit{e.g.}, synchronous or asynchronous) and both \textit{proactive} and \textit{reactive} behavior strategies, in a counterbalanced order. Passive behavior was present in all conditions, as we recorded the full session. Families were randomly assigned one of the conditions.

\begin{table*}[t]
\centering
\caption{Demographic Information of Families who Participated in the Laboratory Study.}
\label{tab:study2-families}

\begin{tabular*}{\textwidth}{@{\extracolsep{\fill}} c c c l l c c c @{}}
\toprule
\textbf{Family Code} & \textbf{Condition} & \textbf{Initial Behavior Strategy} & \textbf{Parent(s)} & \textbf{Child(ren)} & \textbf{Ethnicity} & \textbf{Races} & \textbf{Income} \\
\midrule
A1  & \textit{async.} & Proactive & 42F & 8M & Mixed & B, W & \$100k -- \$150k \\
A2  & \textit{async.} & Proactive & 43M & 12M & NH/L & W & \$200k or more \\
S3  & \textit{sync.}  & Reactive  & 36F & 7M & NH/L & A & \$100k -- \$150k \\
A4  & \textit{async.} & Proactive & 38F, 38M & 9F & NH/L & W & \$150k -- \$200k \\
A5  & \textit{async.} & Reactive  & 47F, 50M & 14F, 10M & NH/L & W & \$75k -- \$100k \\
A6  & \textit{async.} & Reactive  & 39F & 12F, 11F, 8M & NH/L & W & \$75k -- \$100k \\
S7  & \textit{sync.}  & Reactive  & 44F & 11M, 7M & NH/L & W & PNR \\
S8  & \textit{sync.}  & Proactive & 42F, 46M & 9F & NH/L & W & \$150k -- \$200k \\
A9  & \textit{async.} & Reactive  & 35M, 37F & 10F, 7M & NH/L & A & \$200k or more \\
A10 & \textit{async.} & Reactive  & 41M & 7F & NH/L & A & \$200k or more \\
S11 & \textit{sync.}  & Reactive  & 49F & 12M & NH/L & W & \$100k -- \$150k \\
S12 & \textit{sync.}  & Proactive & 41F, 50M & 5M & NH/L & A & \$150k -- \$200k \\
S13 & \textit{sync.}  & Proactive & 40M & 12M & H/L & AIAN, W & \$100k -- \$150k \\
A14 & \textit{async.} & Proactive & 41M & 12M & NH/L & W & \$150k -- \$200k \\
S15 & \textit{sync.}  & Proactive & 43F & 10M & NH/L & W & \$150k -- \$200k \\
S16 & \textit{sync.}  & Reactive  & 50F, 55M & 12F & NH/L & A & \$200k or more \\
S17 & \textit{sync.}  & Proactive & 44F & 11M, 9F & NH/L & W & \$200k or more \\
A18 & \textit{async.} & Proactive & 49M & 11M & NH/L & W & \$100k -- \$150k \\
S19 & \textit{sync.}  & Reactive  & 37F & 8M, 4M & NH/L & W & \$200k or more \\
A20 & \textit{async.} & Reactive  & 39F & 7F & Mixed & W & \$200k or more \\
\bottomrule
\end{tabular*}

\vspace{2pt}
{\raggedright\footnotesize\textit{Note.} Abbreviations are used. \textbf{Races:} W = White, B = Black or African American, A = Asian, AIAN = American Indian or Alaska Native. \textbf{Ethnicity labels:} ``Mixed'' indicates both Hispanic/Latino and Non-Hispanic/Latino family members; NH/L = Non-Hispanic/Latino; H/L = Hispanic/Latino. \textbf{Other:} PNR = Prefer not to respond; \$100k = \$100{,}000.\par}
\end{table*}

\subsection{Study Procedure}
% Describe the room setup
The study took place in a lab environment where two rooms were connected via a hallway, designed to look and feel like natural living spaces. One has a more adult-oriented layout simulating a living room, in which we placed a tablet -- we will refer to this as the \textit{Tablet Room}. The other room had chairs and tables meant for children, and is where the robot is placed -- we will refer to this room as the \textit{Robot Room}. Each room is adjacent to an observation booth equipped with two-way mirrors, and monitors with access to live video streams from discrete cameras installed in the lab spaces.

The lab session lasted a total of 75-90 minutes, including the introduction and consent process (10 minutes), a warm-up and settling down period (5 minutes), the first condition (20 minutes), a short break (5 minutes), the second condition (20 minutes), an opportunity for families to debrief each other and wrap up (5 minutes), and finally a semi-structured interview, a brief demographics survey, as well as payment handling (15 minutes).

At the end of the introduction phase, families decided how they would like to split up between the two rooms, and also decided on what activities they would like to do. Families were encouraged to bring their own activities, but were also provided with playful activities by the researchers. As fallback options, in the \textit{Tablet Room}, we prepared a variety of magazines and small games (\textit{e.g.}, a puzzle and a vertical 4-in-a-row game board). In the \textit{Robot Room}, we prepared the same marble run kit we've used in the technology probe sessions (Section~\ref{sec:design_techprobe}). Family members were encouraged to stick with their choice of room within a condition, but they were welcome to switch rooms between conditions. 
% Families were then given five minutes to familiarize with the rooms and the environment, at the end of which, the experimenters informed participants in both rooms that the 20-minute session for that condition has started, as they closed the doors to the rooms.

During each 20-minute session, the experimenters sat in the observation booth adjacent to the \textit{Robot Room}, and controlled the robot following a Wizard-of-Oz protocol: % in a WoZ fashion. The general protocol for robot movement was as follows. 
The robot starts at the corner of the room, and at minutes 2, 9, and 16, it would navigate to where the activity is taking place, near one of the participants. After the interaction (either a video recording or a video call), or the absence of any interactions after a few minutes, the robot would navigate back to its resting position in the corner. This return to the corner is sometimes skipped if users continued to interact with the robot.
% If a user interacts with the robot (\textit{e.g.}, pressing a button on the screen, moving the robot, adjusting the head tilt, etc.), the experimenter will delay any manual movements, sometimes skipping the return to the resting position altogether if there are frequent interactions during the session.
% After the first 20-minute session, the experimenters would open the doors, and remind the participants that they have a 5-minute break before the next 20-minute session.
The main task of the experimenters was to control the robot's movement through a combination of pre-configured locations and joystick-based manual navigation. In the synchronous condition, if the call is still active at the 3-minute mark, the experimenter would manually end the call (with a warning at 2:45). More details on the system setup is available in Supplementary Materials.

% the experimenters also had to occasionally handle the simulation of ending a call (by muting and hiding the video call windows). 
% if a call reached 2 minutes and 45 seconds, the experimenter would trigger a warning message to both the tablet and the robot, reminding them of the 3-minute limit.

Between the 20-minute sessions, participants had a 5-minute break, and were allowed to walk around. After the second session, families had another 5 minutes to exchange experiences and thoughts, before the experimenters started the semi-structured interview. The semi-structured interviews focused on the family's general experience during the session, their thoughts on the robot behaviors and the two different strategies (proactive, reactive), the video captures or calls, and how they envision a system like this could (or could not) fit in their family to facilitate connection makings. The session ended with a brief demographics survey, and a payment of \$30 USD provided for each family.

\begin{table*}[t]
\small
  \centering
  \caption{Template analysis summary for the user study. The table summarizes themes identified across families, for each theme, the number of families in the synchronous condition, the asynchronous condition, and the total across both conditions.}
  % \vspace{-10pt}
  \label{tab:thematic}
  \setlength{\tabcolsep}{2pt}
  \begin{tabular*}{\textwidth}{@{\extracolsep{\fill}} p{4.5in} c c c @{}}
    \toprule
    \textbf{Theme} & \textbf{Synchronous} & \textbf{Asynchronous} & \textbf{Total} \\
    \midrule
    
    \textbf{Robot-Facilitated Connection} &  &  &  \\
    Value of hands-free interaction. & 8 & 3 & 11 \\
    Factor of convenience. & 3 & 4 & 7 \\
    Felt closer to their family member on the other end. & 3 & 3 & 6 \\
    Technology-mediated connection still falls short. & 5 & 1 & 6 \\
    \midrule
    
    \textbf{Differed by Modes of Communication} &  &  &  \\
    Monitoring or supervision use case. & 4 & 10 & 14 \\
    Privacy considerations. & 2 & 6 & 8 \\
    Physical embodiment \& the afforded control, elevated sense of an additional \textit{being} in the room. & 6 & 0 & 6 \\
    \textbf{Direct Comparisons} &  &  &  \\
    \textit{Sync.} enables easier \textit{two-way} communication. & 5 & 4 & 9 \\
    \textit{Sync.} offers real-time communication. & 5 & 2 & 7 \\
    \textit{Sync.} is more interactive. & 3 & 3 & 6 \\
    \textit{Sync.} requires both to be available at the same time. & 4 & 4 & 8 \\
    \textit{Async.} can contribute to digital memory. & 2 & 5 & 7 \\
    \midrule

    \textbf{Robot's Behavior Strategies} &  &  &  \\
    Reactive strategy afforded families greater autonomy and control.  & 5 & 6 & 11 \\
    Proactive strategy was disruptive. & 5 & 0 & 5 \\
    Proactive strategy made the robot more autonomous \& allowed users to focus on the on-going activity. & 1 & 4 & 5 \\
    Proactive strategy allowed the robot to capture moments that would have otherwise been missed (candid, spontaneous, or fleeting moments). & 0 & 7 & 7 \\
    Proactive actions can serve as reminders to connect. & 3 & 1 & 4 \\
    \bottomrule
  \end{tabular*}
  \setlength{\tabcolsep}{6pt}
  % \vspace{-12pt}
\end{table*}

\subsection{Data Collection and Analysis}
Both the \textit{Tablet Room} and the \textit{Robot Room} were video recorded with multiple discrete cameras. Captured video clips (asynchronous) and screen recordings of video calls (synchronous) were saved, and interviews were audio recorded and transcribed. The first author conducted an initial template analysis \cite{brooks2015utility}, developing a preliminary codebook based on the full set of interview transcripts. A second coder was then trained and completed an independent round of coding using this preliminary codebook as a starting point. The first author and second coder subsequently met to discuss and iteratively resolve differences, refining the codebook and the set of themes. These themes were further discussed and refined through conversation with the broader research team. We compared themes across the two study conditions. Video clips and recordings were used as complementary data sources to contextualize interpretation of the interview findings. When reporting findings, we refer to families in the asynchronous condition with a prefix \textbf{A}, and those in the synchronous condition with a prefix \textbf{S} (\textit{e.g.}, A1 \& S3).

\section{Findings}
% \todo{There are some sections in the results that I feel were over-explained, had many filler sentences, or that I worry readers might have trouble following. I tried flagging them, but I’d love to hear what others feel once they make a pass on the paper.}

\label{sec:findings}
% Our thematic analysis yielded a range of insights. In what follows, we discuss 
Our thematic findings reveal how families perceived the robot's effectiveness in facilitating communication, how synchronous and asynchronous communication shaped these perceptions, and how proactive and reactive behavior modes were received (see Table~\ref{tab:thematic}). %We also summarize the types of opportune moments that participants recommended for robot-facilitated sharing and connecting with family members. 

\subsection*{Impact on Communication and Connection}
18 out of the 20 families (all but A1 and A9) recognized some aspects of the robot's potential in facilitating communication and connection-making. For example, the son in S3 (sync) described the feeling as ``\textit{It's like I have connection with her (the mother).}'' The system enabled a convenient and easy way of communication, mainly due to its autonomous and mobile aspects, as described by the mother from S8: ``\textit{[the robot] could just be moving around and then be able to see what we were up to, without [us] having to change anything.}'' In a similar vein, but comparing the robot with a phone call, the father of A4 described how with the robot, he wouldn't need to ``\textit{go and answer the phone}'', that the robot ``\textit{would just come to me.}'' The proximity of the robot also created a sense of convenience that encouraged opportunities for connection: ``\textit{... if the robot's rolling around, I'm probably going to be like, yeah, I'm recording myself playing [music]. It's right there}'' (daughter, A5). 11 families (eight from synchronous, and three from asynchronous) specifically appreciated the hands-free nature of the robotic system. Recounting her experience video-calling her parents in a foreign country to show them her then newborn daughter, the mother from S16 said that instead of holding a phone, ``\textit{with the robot, it would be much, much easier to reach that goal, to have my family far away see what's going on on our side.}'' In addition, some users described experiences of heightened presence when communicating through the system (A2, A6, A10, S11, S15, S16). As one child explained, ``\textit{Especially when I could call my mom,  it made me feel like the robot is my mom. [...] it did feel like you were right there}'' (son, S11). Beyond their own households, participants envisioned other useful scenarios for: older or less mobile family members (eleven families), relatives living far away (eight families), young children (seven families), and family members who were traveling (five families). Some families also imagined using the robot to check in on pets (six families) or to connect across different floors in multi-story homes (four families). These reflections suggest broader potential for supporting connection-making across diverse family contexts.

Although most participants saw the potential and value in a robotic platform, some of them had reservations. In addition to some limitations of our study setup (\textit{e.g.}, video viewing experience on the devices, quality of the video call, and ease of controlling the robot), these reservations were often regarding technology-mediated communications in general (and not specific to robots as the mediator). As the mother of A6 summarizes, ``\textit{... there are just some things that don't translate as easily through the screen.}'' The father of S13 lay it out in more detail: ``\textit{For me it's about proximity. When I walked over there to him [during the break], it's like being close, seeing what he's doing. I'm able to read his facial expressions more. On video it's very muted. We always put on a face when we're doing video calls for some reason. We don't fully communicate.}''

\subsection*{Modes of Communication}
% Synchronous vs. Asynchronous}
% To understand how the conditions impacted the families' experiences and perceptions, we report on themes that differ noticeably across the conditions, as well as a summary of explicit comparisons discussed by the participants.
% First of all, a
Families recognized the benefits of the \textit{monitoring and supervision} aspect of the system, comparing it to a security camera or a \textit{baby monitor} (all async families and S3, S7, S11, S17); discussed \textit{privacy concerns} (6 in async, 2 in sync); appreciated and anthropomorphized the \textit{physical presence} of the robot, elevating the sense of having an additional \textit{``being''} in the room (sync condition: S7, S8, S11, S15, S16, S17).
%On the contrary, only four families in the synchronous condition (S3, S7, S11, S17) talked about this aspect. On a related topic, six families in the asynchronous condition discussed considerations about their privacy, while only two in the synchronous condition did. In terms of the \textit{perception of the robot}, six families in the synchronous condition (S7, S8, S11, S15, S16, S17) mentioned that the physical embodiment of the robot and the control afforded by the embodiment, elevated the sense of having an additional \textit{``being''} in the room. 
For some, this meant that the robot felt more than a tool. ``\textit{It seems like there's some sort of a being there that's watching you and interpreting what you're doing and assessing you,}'' explained the father of S8, ``\textit{versus a phone, which is basically doing the same thing, but it doesn't feel that way.}'' For others, the presence of the robot brought the remote family members closer, revealing effects of anthropomorphism `\textit{Because you control the robot, so the robot becomes you once you get to take it over.}''
%Talking about how he felt about his mother controlling the robot during the call, the child from S17 said to his mother: ``\textit{Because you control the robot, so the robot becomes you once you get to take it over.}''

%During discussions about how synchronous and asynchronous communications may differ in facilitating connection-making, families generally agreed that each could be useful under different scenarios. 
Participants often weighed the usefulness of the synchronous and asynchronous communication modalities. Nine families reflected on how synchronous communication may better support \textit{two-way} communication. Similarly, seven families noted that video calls better afford \textit{real-time} exchange of information, offering a more interactive experience (A4, A6, S8, A14, S15, S17). On the other hand, video calls were perceived as more demanding of users’ attention, requiring both parties to be \textit{in the moment}, whereas with asynchronous communication, the receiving person does not necessarily need to be available (A2, A4, A6, S8, S12, S16, S17, A20). An additional benefit of asynchronous communication was that video clips could contribute to the family's digital memory (A2, A5, A6, A10, S11, A14, S16): ``\textit{... over time, looking at videos from this year, to the next year, to the next year, and being able to be like, oh, they grew so much,' or they changed', or `now they're able to play a whole song of music as compared to just a couple notes'}'' (Father, A14).

\subsection*{Robot's Behavior Strategies}

All participants experienced both proactive and reactive robot behaviors in counterbalanced order. Across both synchronous and asynchronous conditions, a common theme was that the \textit{reactive strategy afforded families greater autonomy and control over when communication occurred} (11 families). This included both preventing the robot from acting when it was not desired (\textit{e.g.}, ``\textit{what if you don't want a video?}'' -- son, A1) and initiating interaction on demand (\textit{e.g.}, ``\textit{[the reactive mode] is better because I can control when to contact him.}'' -- mother, S3). This resonates with how the families in the technology probe sessions preferred a mostly passive robot that blended into the background for most of the time. In addition, several families noted that such control enabled them to set up interactions more intentionally, creating space for more purposeful communications (A2, S12, S15). In contrast, five families in the synchronous condition explicitly expressed concerns that the proactive mode was disruptive and could be annoying (S3, S7, S11, S12, S13). Other families echoed a broader sense of uncertainty associated with the proactive strategy, describing a feeling of needing to remain on stand-by to interact with the robot (S3, A6, S17).

Families in the asynchronous condition, on the other hand, were comparatively more willing to relinquish some of that control and described several benefits of the proactive mode. First, some valued the proactive strategy because it made the robot feel more autonomous, reduced the need for user attention, and allowed them to stay focused on ongoing tasks (A9, A10, S11, A14, A20). Second, participants appreciated how proactive behavior enabled the robot to capture and share candid or spontaneous moments, which could sometimes feel more meaningful than deliberately staged clips (A1, A2, A6, A20). As one father (A2) put it: ``\textit{... it would be cute to have just a random video of your kid reading a book on the couch, as opposed to, `Hey look, I finished the book.'}'' Finally, families also highlighted that the proactive mode could capture fleeting, everyday moments that might otherwise be overlooked (A9, A14, A18), and could serve as a prompt for families to communicate and connect when they might otherwise forget (S11, S13, A14, S15).

In sum, families generally preferred having greater control over when and how interactions occurred, but they also acknowledged scenarios in which proactive behavior could be advantageous, such as capturing spontaneous moments or prompting communication that might not otherwise take place.

\subsection*{Opportune Moments for Sharing and Connecting}
% When asked to describe potential scenarios for robot-facilitated connection, families identified a wide range of ``opportune moments'' that went beyond simple availability. These moments generally fell into three distinct categories: significant milestones, safety-critical situations, and spontaneous everyday experiences.
Families described opportune moments as: milestones, safety-critical situations, and spontaneous everyday experiences.

\subsubsection*{Milestones and Achievements} 
% The most frequently cited use case was the sharing of special events and achievements. 
Ten families (four \textit{sync}: S8, S11, S16, S17, six \textit{async}: A5, A6, A10, A14, A18, A20) envisioned the robot capturing or prompting a connection during special occasions such as birthday parties. Similarly, nine families (four \textit{sync}: S15, S16, S17, S19, five \textit{async}: A5, A6, A10, A14, A20) discussed using the robot to showcase ``cool things'' or skills, with specific examples including playing a musical instrument, doing gymnastic moves, or completing a puzzle. In these instances, the robot was viewed as a way to validate and cherish the child's accomplishments and to also share them with remote family members who would otherwise miss these important moments.

\subsubsection*{Emergencies and Pragmatics} 
Eight families (five \textit{sync}: S7, S8, S12, S13, S16; three \textit{async}: A6, A10, A14) mentioned the robot's potential role in emergencies, such as alerting parents if a child was ``distressed'' (Father, A10). Some families also noted more pragmatic uses, such as checking in on pets (4 families) or monitoring homework progress (3 families).

% Eight families (five \textit{sync}: S7, S8, S12, S13, S16, three \textit{async}: A6, A10, A14) mentioned the robot's potential role during emergencies, such as alerting parents if a child is ``distressed'' (Father, A10).
% % As the father of A10 puts it, ``\textit{when they're playing in the toy rooms... I would want it to alert me if the kid is distressed. That would be super useful. In our [study's] setting, I felt like having access to her was nice, to make sure that she's okay.}''
% % Notably, this use case was more prevalent among families in the synchronous condition (5 families) compared to the asynchronous condition (3 families), likely reflecting the value of real-time communication in urgent situations. 
% Some families noted pragmatic uses, such as checking in on pets (4 families) or monitoring homework progress (3 families).

\subsubsection*{Spontaneous and Chaotic Moments} 
% Families highlighted the value of the robot in capturing the messy family life. 
Six families (four \textit{async}: A2, A4, A9, A20, two \textit{sync}: S7, S19) explicitly mentioned ``fun,'' ``silly,'' or ``chaotic'' random moments as ideal targets for the robot to capture. %Families in the asynchronous condition preferred spontaneous capture more than the synchronous condition.
 % Interestingly, this desire to capture spontaneity was twice as common in the asynchronous condition (4 families) as in the synchronous condition (2 families). 
 % This suggests that while emergencies may demand a live call, the silly and ephemeral moments of daily life are seen as perfect content for asynchronous sharing. 
 % This also contributes to the ``digital memory'' of the family by allowing busy and otherwise unavailable parents to witness the texture of daily life retrospectively. 
 % As the mother from A4 explained, the robot could enable her to ``\textit{... come back and see a whole bunch of things that have occurred at random throughout the day}'' even if she was ``\textit{booked from like 8 to 8}'' and unable to step away for a synchronous interaction.

\section{Discussion}
% Let's see how we do in space -- we could just focus on design implications if tight on space.
% Our findings revealed how families perceived the robot's potential to support connection-making through distinct roles in facilitating synchronous and asynchronous communication, the trade-offs between proactive and reactive robot behaviors, and the variety of potential family-robot interactions that may arise. We draw out the design implications for future social robots in the home, and reflect on the limitations of the present work, which point to promising areas for future investigations.

Across the two studies, families did not treat the robot merely as another communication device. Instead, they responded to it as an embodied mediator that could reshape \textit{when} connection becomes possible, \textit{how much effort} it requires, and \textit{what} that connection feels like. Its mobility and hands-free presence lowered the burden of checking in, its embodiment sometimes made remote family members feel more present, and its ability to initiate or capture interaction introduced new tensions around interruption and autonomy. Taken together, these findings suggest that the design of family communication robots is not only a matter of supporting message exchange, but of shaping the relational conditions under which family connection is initiated and negotiated.

\subsection{Ethics and Privacy Considerations}
Our findings highlight a critical tension between fostering communication and enabling surveillance, particularly evident in the asynchronous condition where participating families compared the robot to monitoring devices such as a security camera or baby monitor. Families' discussions of monitoring, supervision, and loss of control, especially in proactive configurations, reveal how a system designed to support communication can quickly come to feel like surveillance if not carefully constrained. These tensions were not uniformly perceived across conditions, suggesting that ethical risk is shaped less by the mere presence of a robot and more by how initiative and visibility are distributed across family members and communicated through the robot.

This issue is especially salient in relation to children's privacy and developing autonomy. Families' repeated use of the ``baby monitor'' analogy raises important questions about where the line between parental care and undue surveillance lies for children in this age range. Designing for family connectedness therefore requires guardrails that differentiate between ``checking in'' and ``watching over.'' More broadly, our findings suggest that family robots should be designed not only to protect privacy, but to support \textbf{negotiated visibility}: ways for parents and children to shape what can be seen, when it can be shared, and by whom. In this framing, visibility is treated not as a fixed binary setting, but as something socially negotiated through the robot's design. In practice, this might involve transparent recording cues and accessible mechanisms for children to pause, redirect, or decline observation. Such features could help ensure that the robot's presence feels supportive rather than supervisory, while better respecting the child's emerging autonomy.

\subsection{Parent-Child Interactions}
Consistent with our design concept, our findings suggest that robot-mediated communication reshaped parent-child interactions not by replacing conversation, but by mediating \textit{how} communication was initiated, negotiated, and experienced. Rather than acting as a conversational partner, the robot primarily functioned as an interactional scaffold~\cite{chen2025social} that influenced family dynamics.

\subsubsection*{Robots as Interaction Shapers}
The son in S11's comment that it felt like the ``robot is my mom'' illustrates how the robot was experienced as a proxy for parental presence rather than a competing social actor. This aligns with theories of technological mediation, which suggest that artifacts do not merely transmit information but actively co-shape the relations between human subjects~\cite{verbeek2015cover}. In this way, the robot shaped the interactional conditions of parent-child communication without substituting for its relational content. Its value lay less in what it ``said'' and more in how it enabled hands-free, mobile, and context-rich forms of connection between parent and child. Put differently, the robot contributed not by supplying the substance of parent-child interaction, but by altering how one family member's presence could be projected into another's space.

\subsubsection*{Initiative as Interaction Framing}
The question of ``who initiates the interaction'' was a salient factor in shaping parent-child experience. Reactive configurations afforded families greater agency and intentionality, while proactive behaviors sometimes felt disruptive or imposed. From a relational perspective, initiative functions as a social signal, shaping whether communication feels welcome or intrusive. Moreover, initiation itself also signals an intention to connect: ``\textit{... when it's the user [who initiates] then you know that someone's actually trying to contact [...] it feels like they want to connect with you}'' (son, S13). These findings resonate with theories of autonomy and self-determination~\cite{deci2008self}, suggesting that preserving users’ control over initiation is particularly important in parent-child contexts where power asymmetries already exist, and that the intention behind the explicit communications may be just as important as, if not more than, the content of the communications itself. More broadly, our findings suggest that initiative should be understood as design knowledge rather than simply as a capability. In family contexts, \textbf{who initiates interaction carries social meaning}: preserving user control over initiation can support agency and intentionality, while robot-initiated interaction can sometimes feel intrusive or imposed. Designers should therefore treat initiation not only as a question of functionality, but as a relational choice that shapes how communication is experienced.

\subsubsection*{Robots as Low-Stakes Intermediaries}
The robot also changed the threshold for connection by enabling forms of contact that required less coordination and less immediate reciprocity. Specifically, families valued robot-mediated interactions most when they reduced effort without demanding immediate emotional engagement. Asynchronous communication, in particular, allowed connection to occur without requiring mutual availability, while the robot's presence softened the social friction of initiating contact. In this role, the robot functioned as a ``safe intermediary'' for low-stakes, phatic connection~\cite{vetere2009phatic}, supporting communication while respecting emotional readiness. This suggests that effective designs should prioritize emotional sustainability over maximizing engagement.

\subsubsection*{Playful Appropriation of the Robot}
While interest levels varied, children frequently engaged the robot playfully, incorporating it into games such as hide-and-seek or monkey-in-the-middle (Figure~\ref{fig:playing_with_robot}), or testing its response. Such behaviors reflect children's tendency to make sense of new technologies through play~\cite{ali2021social}, using exploration to negotiate roles and boundaries. Importantly in our context, these playful interactions did not detract from moments of connection. Rather, they coexisted and sometimes even augmented the family connection, for example, when some of the plays involved the family members in the tablet room. These snippets, while anecdotal, further suggest that the robot's contribution may lie not only in transmitting communication, but also in \textbf{creating new occasions for shared interaction} around the robot itself.

Taken together, the robot's contribution was not to replace parent-child communication, but to reconfigure its interactional ecology: who feels present, who initiates, how demanding connection feels, and what new forms of shared activity become possible.

% Design implications
% (both for those explicitly stated and those synthesized from other categories)
\subsection{Design Opportunities}
Families discussed a wide range of preferences on robot behavior strategies and communication modes that very often depend on a hierarchy of contextual factors. We discuss five practical design recommendations based on the findings.

% Our findings in Section~\ref{sec:findings} directly point to several design implications. For example, the families' wide range of responses about the preferred robot behavior strategy underscores the importance of carefully \textit{balancing robot autonomy and user control and implementing context-sensitive behavior strategies}. Similarly, the diversity of preferences regarding synchronous and  asynchronous communication highlights the need for \textit{systems that flexibly support both modes and enable families to move fluidly between them}. We discuss four additional implications.

\subsubsection*{Recognize Opportune Moments for Connection-Making}
% \todo{Focus on WHEN to do something}
Since each behavior strategy and mode of communication has its own use case, simply offering families an extensive menu of options may not be sufficient or ideal. Families envisioned greater value in a robot that could recognize \textit{when} to act: \textbf{identifying opportune moments for connection and autonomously take action, either by capturing a moment or initiating a call}. Prior work in HCI and HRI has examined opportune moments for interaction, but primarily from a time-based perspective (\textit{e.g.}, day and time)~\cite{hsu2024now, de2016long, lopatovska2019talk}. In our context about moments for connection-making, time-based rules may not be sufficient. Instead, families discussed with us a wide range of potential moments, including \textit{special events} such as birthday parties, \textit{notable achievements} like playing a musical instrument, \textit{emergency} situations, \textit{creations} such as a finished puzzle, and fun, silly, or even chaotic \textit{snippets of everyday life}. Leveraging the robot's unique ability to proactively detect and respond to these opportunities may significantly enhance its role in facilitating \textit{well-timed, quality} connections within the family.

At the same time, our study setup does not allow us to fully disentangle whether negative reactions to proactive interaction were driven by robot initiative itself or by how and when that initiative was introduced. Because proactive interactions in the laboratory study were delivered through a standardized WoZ protocol, participants responses may have reflected both who initiated and whether the robot entered attention at an inopportune moment. Some participant comments point to timing as a likely contributing factor. As the mother from S7 explained, ``I preferred when we called, because we were just starting to read the book, and then it was calling. It was kind of an interruption.'' Comment like this do not fully resolve the attribution issue, but they suggest that the perceived disruption of proactive interaction may have stemmed at least in part from poor timing. This further highlights the importance of recognizing not only meaningful moments for connection-making, but also \textit{moments when interruption is socially appropriate}.

% This is also consistent with preferences voiced by families in the Formative Study, who assessed positively a robot that is passively observing for most of the time (and perhaps curating a list of captured moments that it may share with the family later).

\subsubsection*{Adjust Behavior Strategy and Mode of Communication Based on the Context}
Once an opportune moment is recognized, the robot must determine \textit{what} to do and \textit{how} to do it. Consistent with prior work~\cite{ju2008design, fink2014robot, chevalier2022context}, we argue that \textbf{the robot should select both a behavior strategy and a mode of communication based on situational context}. Relevant factors may include \textit{the nature of the moment, its urgency, the availability and interruptibility of family members, and the family's existing communication routines}. In our findings, the desire to capture spontaneity was twice as common in the asynchronous condition (four families) as in the synchronous condition (two families), suggesting that while emergencies may call for live communication, silly and ephemeral moments of daily life may be better suited for asynchronous sharing. In this way, asynchronous clips can contribute to the family’s “digital memory,” allowing otherwise busy or unavailable parents to witness the texture of everyday life retrospectively.

\textbf{Temporal contexts} could also be taken into account: \textit{When was the last time these family members connected?} While we do not have a readily available flow chart that would work out-of-the-box for all relevant systems, these general considerations may help guide the robot in making a decision. For example, imagine a robot notices that a child is playing piano. Urgency is perhaps low, so asynchronous communication may be sufficient --- unless this is happening on a Friday evening, during which the family would typically make their weekly call to the grandparents, then perhaps the robot could consider synchronous communication as well. \textit{Are the parents home? Are they attending to other tasks that require their focus? Should the robot attempt a co-located form of synchronous communication, by inviting the parents to the piano practice?} 

Although with recent advances in Vision Language Models, robust perception and comprehension of context is becoming increasingly feasible~\cite{fang2025mirai,lim2024exploring,xu2026designing}, these decisions are unlikely to generalize uniformly across families. Families differ substantially in interruption tolerance, privacy preferences, and daily rhythms, suggesting that a deployable system would likely require \textbf{a personalization pathway rather than a one-size-fits-all strategy}. One feasible direction would be for the robot to begin with lightweight configuration and conservative defaults, and then gradually adapt through family feedback during use. For example, families might indicate when prompts are helpful or ill-timed, how often the robot should initiate interaction, and under what conditions asynchronous sharing should escalate to synchronous communication. \textbf{Over longer-term deployments, such adaptation could become increasingly fine-grained through online preference learning or co-adaptation}, allowing the robot's context-sensitive behavior to better align with a particular family's routines and norms.

% It is, however, critical to recognize that these contextual decisions are not age-neutral. Our findings hint that \textbf{the developmental stage of the child may also shape how robot behaviors are interpreted}: younger children were more likely to engage the robot playfully or treat it as a proxy for parental presence, while older children demonstrated a heightened sensitivity to unsolicited capture and monitoring. This highlights a need for \textbf{age-adaptive behavior strategies}. We propose a \textbf{developmentally-aware trajectory model for HRI}: as children transition from early childhood to adolescence, the robot's default behavior strategy should undergo a structured shift from proactive or scaffolding roles to passive or autonomy-supportive roles. For younger children, proactive behaviors were often interpreted through a lens of play; however, for older children, the same initiative risks being perceived as infantilizing or as a form of parental surveillance. By designing robots that ``grow'' with the child, gradually ceding initiative to the user as they become developmentally more mature, designers can ensure the system remains a supportive interactional scaffold rather than a source of relational friction.

\subsubsection*{Reduce User Effort in Establishing and Maintaining the Connection}
% \todo{Lower-level HOW to do}
Once the robot has determined a behavior strategy and mode of communication, it must also decide how to carry out its plan in ways that minimize user effort and burden. Families in our study especially appreciated hands-free interactions and the robot's mobility, echoing prior findings on the importance of ease of use for technology and robot acceptance~\cite{adams1992perceived,de2019would}. In this context, we argue that \textbf{interactions should be designed to lower the effort required of families to establish and maintain the connection}. Concretely, this translates into features such as \textit{autonomous navigation} to suitable positions, \textit{adjusting the camera angle} to capture the activity of interest, and automatically \textit{zooming, tilting, or panning} when needed. From a human-robot interaction perspective, additional modalities such as natural language commands or gestures may also be useful to keep the human in the loop when necessary~\cite{elazzazi2022natural,nwankwo2024conversation}. These considerations highlight not only the unique advantages of a robotic platform, but also their particular relevance for populations that families identified as key beneficiaries of such a system, such as children and older adults.

% Families' appreciation for hands-free interactions and the robot's mobility underscores the value of designs that reduce user effort. Six families (A4, S7, A9, S11, S16, S17) mentioned how voice commands or other forms of voice-based initiation would make it easier to engage with the robot (and, by extension, their family). Three families in the synchronous condition (S8, S12, S13) further noted that simplifying control of the robot, such as automatically focusing on nearby activities, would reduce the burden of managing interactions. These considerations are particularly salient for some populations that families envisioned as key beneficiaries of such a system, such as the elderly and the young. As the father of F13 explained, ``\textit{If it's a little kid or an older person, can they easily move it and show what they want to show? Can they easily control it? Those would be a couple of things that I think of [in terms of] accessibility.}'' This feedback, emphasizing convenience and reduced effort in communication, also highlights the unique advantage of robotic platforms: their ability to autonomously and adaptively facilitate interactions in ways that may be challenging for other technologies.

\subsubsection*{Anticipate Unexpected Family-Robot Engagements}
\begin{figure*}[tb]
  \centering
  \includegraphics[width=\linewidth]{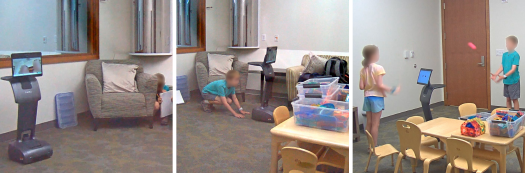}
  \caption{\textbf{Examples of unexpected playful interactions between the robot and the children.} In S17, children attempted to engage the robot in games such as hide-and-seek and monkey-in-the-middle. During a video call with his mother and sister, the son initiated a game of hide-and-seek with the robot, which was being remotely controlled by his family. Later, when his sister joined him in the robot room, the two attempted to play monkey-in-the-middle during periods without active calls, leveraging the robot's built-in ``following'' feature.}
  \Description[Three scenes of children playing with the robot]{Composite figure with three photographs showing playful child-robot interaction in a furnished playroom (one of the lab rooms). In the left image, a mobile robot with a screen stands near an armchair while a child crouches partially hidden beside the chair, suggesting a hiding game. In the middle image, a child crouches low on the floor facing the robot, which is positioned nearby in the open room. In the right image, two children stand on opposite sides of the robot and toss a ball across the room, with the robot placed between them near a child-sized table and toy bins. Across the three images, the robot is treated as a social play partner rather than only a communication device.}
  \label{fig:playing_with_robot}
  % \vspace{-12pt}
\end{figure*}

The first three opportunities focus on speculative interaction scenarios, yet unexpected engagements are inevitable. In our study, we observed family members moving or pushing the robot, and children attempting to engage with it in playful ways (see Figure~\ref{fig:playing_with_robot}). For a system to respond appropriately, it must first recognize that it is in such a situation~\cite{bremers2023using}, and then generate a strategy to proceed. Although the specific strategy will depend on the scenario and implementation, we argue that \textbf{some form of ``back-up plan'' is crucial}. Existing work is sparse, but \citet{honig2021expect}'s approach in leveraging the human-robot ecosystem the robot is embedded within, and proactively asking for clarifications or assistance from the family members around, may prove promising. For example, if the robot is blocked or moved abruptly, it could gracefully pause its tasks and request clarification. Such capability is essential not only for robustness but also for fostering trust~\cite{giorgi2022friendly,kok2020trust}, as families will inevitably experiment with and engage with the robot in ways designers cannot fully predict.
  
% Although our study primarily focused on the robot's role in facilitating communication and connection between family members, several observations underscored the importance of direct interactions between the robot and individual family members. For instance, multiple participants (A2, A4, A10, S11, S13) in the \textit{Robot Room} repositioned the robot closer to the activity, even during proactive sessions when they could not otherwise initiate interaction. As one daughter (A10) explained, ``\textit{I was doing it so it could take a video easier, like a little closer}.'' Such moments highlight that robots designed for family connection-making must anticipate and be adaptable to diverse modes of engagement, especially with the children, and account for how their presence, embodiment, and interactivity may influence their usage.

\subsubsection*{Account for the Evolving Family Ecosystem}
Facilitating family connections differs from many other child-robot interaction contexts in that interactions are embedded within a broader family ecosystem that is already complex without the robot~\cite{cagiltay2024toward}. Beyond static preferences, this ecosystem includes evolving relationships, roles, routines, and developmental trajectories. This complexity is consistent with prior work showing that families do not view robot support as uniformly desirable across routines~\cite{xu2024robots}, underscoring the need for family robots to account for heterogeneous expectations even within the same household. Therefore, designing for the family means designing for a moving target.

For robots intended to facilitate family connection, we envision systems that can adapt to the state of connections among family members, incorporating modes, styles, and patterns of communication into decisions about behavior strategy. In our study, children's interest in communicating with parents varied widely, ranging from declining most requests to share captured video to repeatedly initiating contact. Such variation reflects not only individual differences but also \textbf{children's developmental stage}, including their evolving autonomy, privacy expectations, and capacity for emotional regulation. Prior research similarly shows that as children grow older, they increasingly seek autonomy and privacy in their interactions with technology, often resisting systems perceived as parental surveillance~\cite{livingstone2007gradations}. Consistent with this, \citet{shin2021designing} note that many parent-child technologies fail when they assume a static child role rather than designing for changing expectations and relationships over time. Our findings point in a similar direction: younger children were more likely to engage the robot playfully or interpret it as a proxy for parental presence, whereas older children appeared more sensitive to unsolicited capture and monitoring. Consequently, effective designs must adapt not only to family-level routines but also to developmental change. \textbf{Rather than treating the child as a fixed user, family robots should be designed for developmental trajectories}, supporting more playful and protective roles in early childhood while gradually shifting toward privacy-respecting, autonomy-supportive roles as children mature. In this sense, robots that support family connection may need to ``grow'' with the child, gradually ceding initiative as the child's capacities and expectations evolve.

% Families also raised broader considerations about how such a system might fit into their household. In comparing the robot-augmented system with familiar technologies (\textit{e.g.}, phones, tablets, security cameras), several families (S8, A9, S13, S16, A18, A20) acknowledged advantages such as mobility and presence, but questioned whether these benefits were substantial enough to offset the added cost or space of another household device. Others speculated about alternative or complementary uses that may elevate the value proposition of such a robot, including monitoring pets or serving as a general household assistant (A5, A9, A10, S11, S16). Echoing the findings from \citet{cagiltay2024toward}, these reflections underscore the need to design not only for connection-making scenarios but also for the broader ecosystem of technologies, practices, and constraints that shape everyday family life.

% Extending to Specific Populations and Scenarios
% maybe for future work

\subsection{Limitations and Future Work}
Our study has several limitations, which also suggest directions for future research. \textit{First}, the design of our in-lab sessions focused on a setup in which family members were separated into two rooms. While this arrangement allowed us to systematically examine the two key dimensions of robot-facilitated communication identified in the technology probe sessions, it also represents an abstracted scenario that may be not ecologically valid. In everyday life, these dimensions may play out differently --- for instance, when family members are in the same room but engaged in separate activities, or when they are collaborating on a shared activity as in the setup of the in-home technology probes. Future work could therefore explore how these dynamics unfold in more naturalistic, co-located contexts, where the boundaries between togetherness and separation are less clear-cut. \textit{Second}, our prototype system was limited in terms of its capabilities, as it was designed to support the investigation of the two highlighted dimensions. Anecdotal observations from the in-lab sessions suggest that additional opportunities exist beyond our intended focus. For example, some families spontaneously explored playful uses of the robot both with and through the robot. Such moments highlight the potential of future designs to support emergent forms of play and interaction that may also contribute to connection-making within families. \textit{Finally}, there may be concerns about the generalizability and ecological validity of our findings. Although the technology probe was conducted in-home, the User Study took place in a lab environment over a relatively short interaction period. Participants were predominantly white, and 12 families reported an annual household income of $\$150{,}000$ USD or above. Future studies should therefore examine more diverse populations, in longer-term, naturalistic deployments, to better capture how such systems might be appropriated across different family structures, cultural contexts, and everyday routines. A further limitation is that \textit{connectedness} was not assessed using validated quantitative measures. Because our evidence comes primarily from interviews and observed interactions, the paper focuses on characterizing how families perceived the opportunities and trade-offs of robot-mediated communication. Future work could build on this design space through longer-term studies that pair qualitative findings with established connectedness or relationship measures. Finally, in the proactive condition, interactions were initiated through a WoZ protocol with partially standardized timing, and the robot's approach behavior was also controlled to ensure comparable experiences across families. As a result, our findings should not be interpreted as isolating the effect of robot initiative alone. Participant responses to proactive interaction may also reflect how and when the robot entered attention, including whether that timing aligned with what family members were already doing. Future work should examine proactive and reactive behavior in more ecologically valid settings, for example, by allowing the robot to trigger interactions based on contextual sensing, or learned preferences, so that factors such as initiative and timing can be more cleanly disentangled.

\section{Conclusion}
This work explored how robots can be designed to facilitate parent-child communication and connection-making. Through iterative design sessions and a controlled user study, we examined a design space defined by robot behavior strategy (proactive, reactive, and passive) and mode of communication (synchronous and asynchronous), and analyzed how these dimensions shaped families’ experiences of connection, agency, and privacy. These findings should be understood as evidence of perceived value and design opportunity, rather than as a direct demonstration of improved family connectedness. Our findings show that robots could be effective when they act as interactional mediators, reducing effort and shaping the conditions for communication, and that different behavior strategies and modes of communication may have different application scenarios. At the same time, families’ responses revealed tensions around initiative, emotional demand, and surveillance, highlighting the importance of context-sensitive and adaptive designs to appropriately leverage the system's capabilities in various scenarios. Building on these insights, we articulated design implications that emphasize recognizing opportune moments, supporting autonomy, and accounting for the broader family ecosystem. Together, this work offers guidance for designing robots that complement, rather than displace, parent-child communications and family relationships.

\section*{Selection and Participation of Children}
The study protocol was reviewed and approved by the authors’ Institutional Review Board. Families were recruited through the University of Wisconsin--Madison research recruitment mailing list (faculty and staff members). The eligibility inclusion criteria for the parents were: Fluent in English, 18 years or older, has at least one child aged 5-12 willing to participate. Siblings outside this target age range were also encouraged to participate. A total of six families (13 children) participated in our in-home design sessions, and 20 new families (27 children) participated in the follow-up laboratory study. Parents provided written informed consent for themselves and their children prior to any study activities. In addition, verbal assent was collected from the children participants. For the in-home design sessions, families received \$50 USD upon completion of the study. For the laboratory sessions, families received \$30 USD. During the assent process, children were reminded of their right to withdraw from the study at any point. After the study, families were given time to discuss the study, and to learn more about how their participation can benefit the community.

\begin{acks}
This work was supported by the National Science Foundation award \#2312354. We would like to thank Enhui Zhao for her excellent research assistance. Figures~\ref{fig:teaser} and~\ref{fig:method} incorporate vector art assets by \texttt{macrovector} from Freepik~\cite{freepik}.

% ; Figure~\ref{fig:method} also incorporates Freepik assets not attributed to a specific artist.

% figure 2: freepik and macrovector
% figure 1: macrovector

% Figures~\ref{fig:teaser} and~\ref{fig:example_main_activities} used vector art assets from Freepik~\cite{freepik}. 
\end{acks}

\bibliographystyle{ACM-Reference-Format}
\balance
\bibliography{references}

% \appendix

\end{document}